\documentclass[10pt, a4paper]{article}

\usepackage{lrec-coling2024} 

\usepackage{multirow, booktabs, bbding}
\usepackage[ruled,vlined]{algorithm2e}

\title{LM-Combiner: A Contextual Rewriting Model for Chinese Grammatical Error Correction}

\name{Yixuan Wang$^1$, Baoxin Wang$^{1,2}$, Yijun Liu$^1$, Dayong Wu$^2$, Wanxiang Che$^{1, \ast}$ \thanks{* Email corresponding.}}

\address{$^1$Research Center for Social Computing and Information Retrieval, Harbin Institute of Technology, China\\
$^2$State Key Laboratory of Cognitive Intelligence, iFLYTEK Research, China \\
\{yixuanwang, yijunliu, car\}@ir.hit.edu.cn \\
\{bxwang2, dywu2\}@iflytek.com\\}

\abstract{
Over-correction is a critical problem in Chinese grammatical error correction (CGEC) task.
Recent work using model ensemble methods based on voting can effectively mitigate over-correction and improve the precision of the GEC system.
However, these methods still require the output of several GEC systems and inevitably lead to reduced error recall.
In this light, we propose the LM-Combiner, a rewriting model that can directly modify the over-correction of GEC system outputs without a model ensemble.
Specifically, we train the model on an over-correction dataset constructed through the proposed K-fold cross inference method, which allows it to directly generate filtered sentences by combining the original and the over-corrected text.
In the inference stage, we directly take the original sentences and the output results of other systems as input and then obtain the filtered sentences through LM-Combiner.
Experiments on the FCGEC dataset show that our proposed method effectively alleviates the over-correction of the original system (+18.2 Precision) while ensuring the error recall remains unchanged.
Besides, we find that LM-Combiner still has a good rewriting performance even with small parameters and few training data, and thus can cost-effectively mitigate the over-correction of black-box GEC systems (e.g., ChatGPT).
 \\ \newline  }

\begin{document}
\maketitleabstract

\section{Introduction}
Grammatical error correction (GEC) is a formally simple but challenging task \cite{wang2020comprehensive, bryant2022grammatical}, which aims to identify and correct grammatical errors present in a sentence.
As a basic application task, it has a wide range of applications in areas such as search engines, automatic speech recognition (ASR) systems, and writing assistants \cite{omelianchuk2020gector}.
In terms of model architecture, the mainstream approaches can be categorized into the auto-encoding Seq2Edit model and the auto-regressive Seq2Seq model.

\begin{figure}[t]
\begin{center}
\includegraphics[scale=0.65]{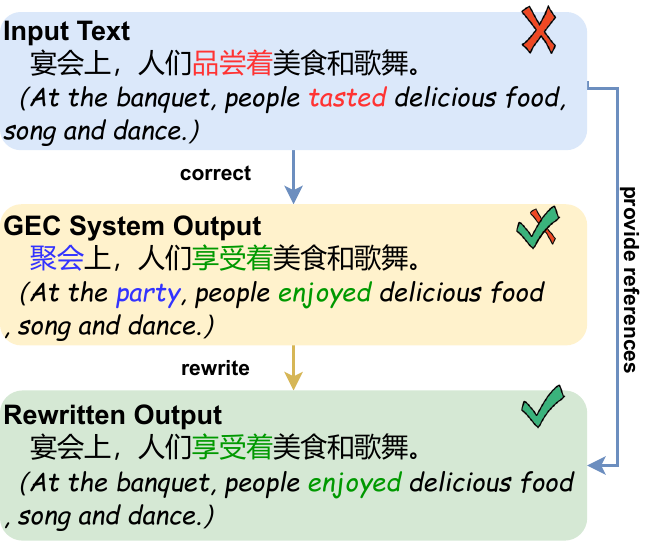}
\caption{
An example of the problem of over-correction,
where \textcolor{red}{red} represents grammatical errors, \textcolor{blue}{blue} represents over-correction, and \textcolor{green}{green} represents correct changes.
LM-Combiner can directly rewrite the system output with reference to the original sentence, filtering out the over-corrections.
}
\label{example}
\end{center}
\end{figure}

Over-correction has always been a challenge in GEC tasks \cite{tang2023pre}, which can seriously affect the precision rate of the GEC system.
As shown in Figure \ref{example}, the over-correction problem is that the error correction system modifies the correct part of a sentence to some other expressions.
Although sometimes these expressions don't differ much from the meaning of the original sentence, as a correction system, excessive modification of the input can still cause annoyance to the user.
Compared to English GEC, Chinese GEC faces a more severe over-correction problem due to the lack of training data and more difficult errors.
Specifically, the previous Chinese GEC task datasets are mainly sourced from non-native learners, with low-quality and inconsistently annotated training sets.
In addition to disfluencies such as spelling errors in English, most of the errors in CGEC involve syntactic and semantic information, which are difficult and make the model prone to false corrections.
The above factors result in the precision of the same baseline model on the Chinese dataset usually being only about half of the rate on the English dataset.
It can be said that over-correction is a key difficulty of CGEC task and deserves a deeper study.

Nowadays, model ensemble is a primary solution to the problem of over-correction.
\citet{li2018hybrid,liang2020bert} view error correction as different types of edit labels and vote to integrate the system based on the labels.
\citet{zhang2022mucgec} integrate multiple architectures of CGEC systems through the method of label voting, improving the precision rate significantly.
\citet{tang2023pre} integrate the outputs of multiple error correction systems at different granularities by computing the perplexity (PPL) through a language model to obtain the final output.
While the above methods can improve the final precision rate, they all suffer from two key problems that need to be solved.
\textbf{(1) \textit{Excessive Cost.}} As ensemble methods, they typically require the results of several models, leading to greater costs in the training phase and longer time in the inference phase.
\textbf{(2) \textit{Reduced Recall.}} Current methods for alleviating over-correction all lead to a significant decrease in error recall rate, which seriously affects the usability of the correction system. 
Voting methods inevitably lead to some decrease in recall, and PPL-based methods can't make accurate judgements on various domain datasets without fine-tuned LMs.

To better mitigate the problem of over-correction, we propose the \textbf{LM-Combiner}, a trainable LM-based text rewriting model.
It can filter the output of a GEC system without a model ensemble, significantly reducing the problem of over-correction while ensuring as much error recall as possible.
In summary, we decouple the over-correction problem from the Chinese grammatical error correction task and treat it as a post-processing rewriting task.
Different from the model ensemble methods, the rewriting model simply takes the original sentence and the result of a single GEC system as inputs, and directly outputs suitable combinations of the two sentences as results.

Specifically, we design the LM-Combiner at the data and model level to ensure its effectiveness.
At the data level, we propose an overcorrected dataset construction method based on the idea of k-fold cross validation.
We divide the training set multiple times, use parts for the model training, and inference on the remaining data to obtain naturally overcorrected sentences.
In addition to this, we propose the gold labels merging approach to further decouple the correction task and the rewriting tasks, so that the LM-Combiner only needs to select from the over-correction and right correction in output sentences of GEC systems.
At the model level, we are inspired by \citet{tang2023pre} to further explore the application of causal language models to the Chinese grammatical error correction task.
Compared to directly using PPL as a criterion, we find that after fine-tuning on the corresponding domain dataset as a rewriting model, GPT2 can better retain the right correction while filtering over-correction,  resulting in higher recall.

We evaluate the proposed method on the FCGEC dataset \cite{xu2022fcgec} sourced from a native speaker corpus.
With the rewriting of the LM-Combiner, we improve the precision of the baseline model by 18.2 points, while ensuring that the recall remain basically unchanged, and the $F_{0.5}$ improves by 5.8 points to reach the level of SOTA.
Besides, experiments show that LM-Combiner has small requirements on model size and data quantity, and can achieve excellent results just by training with base-level models and thousand-level data quantity.

The main contributions of this paper can be summarized as follows:
\begin{itemize}
 \item We propose a novel rewriting model, LM-Combiner, which can effectively mitigate over-correction of the existing GEC systems without model ensemble.
 \item We propose k-fold cross inference, a construction method for over-correction data. It can stably construct over-corrected sentences for LM-Combiner training from existing parallel corpora.
 \item Experiments show that the proposed rewriting method can greatly improve the precision of the GEC system while maintaining the recall constant.
\item We also find that the LM-Combiner achieves good rewriting results even with small parameters and few training data, which provides a cost-saving solution to alleviate the over-correction of existing black-box GEC systems.
\end{itemize}
We will release our code and model\footnote{\href{https://github.com/wyxstriker/LM-Combiner}{https://github.com/wyxstriker/LM-Combiner}}.



\section{Method}

\begin{figure*}[!h]
\begin{center}
 \includegraphics[scale=0.85]{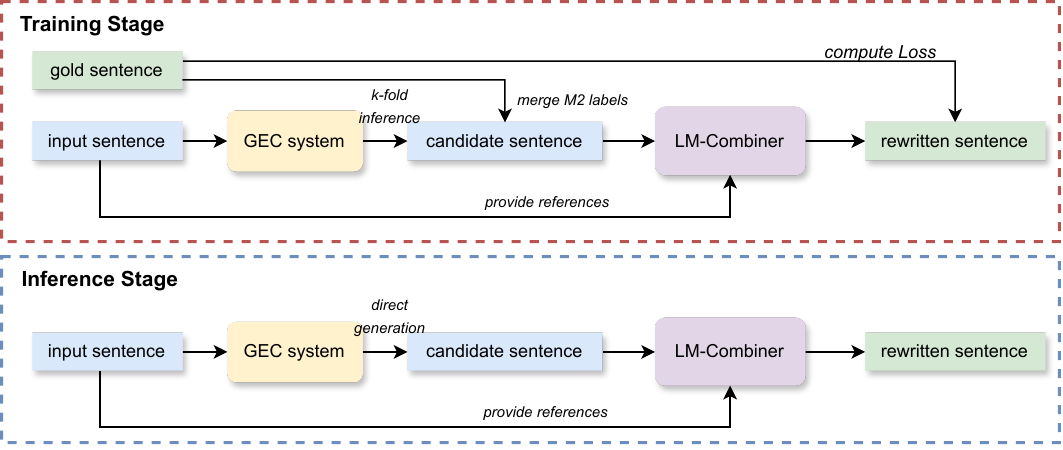} 
\caption{
The flowchart of our error correction-rewriting framework.
In the training phase, we construct candidate sentences containing GEC systems over-correction by k-fold cross inference and gold labels merging (see Section \ref{data_construction} for details).
Then, we train the model to generate gold sentences based on the original and candidate sentences (Section \ref{LM-Combiner}).
In the inference phase, LM-Combiner directly rewrites the system output based on the original sentence.
}
\label{main_fig}
\end{center}
\end{figure*}

The core of our proposed method is a rewriting model, which can alleviate over-correction in the original GEC system by direct rewriting.
The workflow of the correction-rewriting framework is shown in Figure \ref{main_fig}.
Inspired by the recent use of LM for model ensemble \cite{tang2023pre} in the CGEC domain, we train an LM rewriting model that uses only the original sentence and the output of a single system as inputs, which can filter out over-correction and retain as many right correction as possible.
Specifically, we try the application of causal LM on CGEC (Section \ref{baseline}), based on which we propose a rewriting model LM-Combiner (Section \ref{LM-Combiner}) and an over-correction data construction method (Section \ref{data_construction}) for the model training.

\subsection{Causal LM For CGEC} \label{baseline}
The current Seq2Seq-based GEC models are mainly implemented by considering grammatical error correction as a neural machine translation task \cite{junczys2018approaching}. 
Therefore it is natural to use models with encoder-decoder architecture (Bart \cite{lewis2019bart}, T5 \cite{raffel2020exploring}, etc.) to synthesize the capabilities of NLU and NLG for text error detection and correction.
Recently, many causal LM-based models \cite{brown2020language, wei2021finetuned, touvron2023llama} with large-scale corpora and parameters have achieved excellent results on various natural language processing tasks including CGEC. 
It is meaningful to explore the application of relatively small-scale causal LMs like GPT2 \cite{radford2019language} to CGEC task.

For the CGEC task, one of the most obvious ways to use causal LMs is letting the model continue to write the modification result based on the original sentence input.
The inputs to the model during the training phase $S$ can be formulated as:
\begin{equation} \label{gpt_input}
S = \mbox{<sos>} X_1X_2...X_m \mbox{<sep>} Y_1Y_2...Y_n
\end{equation}
where $X$ represents the sentence to be corrected of length $m$ and $Y$ represents the correct sentence of length $n$. <sos> represents the start of generation, and <sep> marks the completion of input and prompts the model to start generating results.
The training labels are obtained by shifting the input as in the traditional LM task, and in order to ensure that the model learns to correct errors, the final training objective of the model is the loss of the correct sentence part, which can be formulated as:
\begin{equation}
\mathcal{L}_{Causal} = 
\sum_{k=i}^{j} -log(P(t_k|t_0t_1...t_{k-1};\theta))
\end{equation}
where $\theta$ is the set of parameters of the language model, $i$ represents the start index of the correct sentence $Y$, $j$ represents the end index of the correct $Y$, and $t_i$ represents the ith token in the model inputs like Equation \ref{gpt_input}.
Although the experiments in table \ref{main_experiment_table} show that the causal LM lacks the ability to correct errors on CGEC compared to the traditional Bart model, its higher precision rate inspires us to employ it as a rewriting model to alleviate the over-correction problem.

\subsection{LM-Combiner Model} \label{LM-Combiner}
\begin{figure*}[t]
\begin{center}
\includegraphics[scale=0.75]{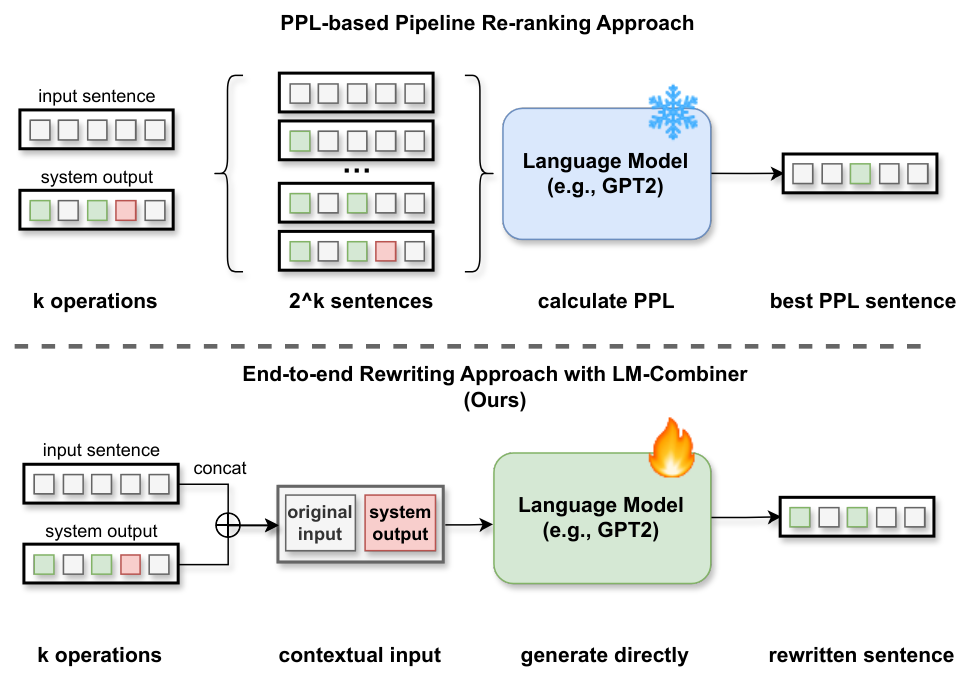} 
\caption{
Comparison between the PPL-based approach and our approach.
Both methods take the original sentence and the output of GEC system as input.
In the figure, \textbf{\textcolor{gray}{gray}} squares represent unmodified tokens, \textbf{\textcolor{green}{green}} squares represent rightly corrected tokens, and \textbf{\textcolor{red}{red}} squares represent overcorrected tokens.
Existing work using PPL to rerank different candidate sentences can improve the precision rate of the system, but the judgment is not accurate enough because the LM is not trained on the domain data, leading to reduced recall.
The LM-Combiner, trained on constructed candidate sentences, is better able to distinguish over-correction and generate results with higher recall end-to-end.
}
\label{compare}
\end{center}
\end{figure*}

Based on the performance of causal LM on the CGEC dataset, we propose the text rewriting model LM-Combiner to deal with the over-correction of the original GEC system.
As shown in figure \ref{compare}, LM-Combiner takes the original sentence and the potentially overcorrected candidate sentences as inputs and directly generates the rewritten sentence as the final output of the GEC system.
The candidate sentences are the outputs of the GEC system, and this method can be regarded as a kind of soft ensemble of the original sentences and the output sentences of a single model.
We first describe the details of LM-Combiner at the model level in this section, and the specific training data construction methods are presented in Section \ref{data_construction}.
Unlike model ensemble methods based on PPL, LM-Combiner is trained to generate rewritten correct sentences directly from contextual inputs (inputs and outputs of the GEC system).
Similar to Section \ref{baseline}, we adopt causal LM as the backbone of our approach.
The inputs to the model $S$ can be formulated as:
\begin{equation}
S = \mbox{<sos>} X_{src} \mbox{<cat>} X_{candi} \mbox{<sep>} Y_{tgt}
\end{equation}
where $X_{src}$ represents the original input sentence, $X_{candi}$ represents the error correction result of the existing model, and $Y_{tgt}$ represents the correct gold sentence. 
The meaning of the special token is the same as in Equation \ref{gpt_input}, and <cat> is used as a split label between the original and candidate sentences.
Like the normal GEC model, in the training phase LM-Combiner only calculates the loss of the correct sentence part, which can be formulated as:
\begin{equation}
\mathcal{L}_{Combiner} = 
\sum_{k=i}^{j} -log(P(t_k|t_0t_1...t_{k-1};\theta))
\end{equation}
where $\theta$ is the set of parameters of the language model, $i$ and $j$ are the start and end indices of the sentence $Y_{tgt}$.

The structure of the LM-Combiner is relatively simple and straightforward, and the key to this model's performance is the way in which the sentence $X_{candi}$ is obtained during the training phase.
The method works only if the $X_{candi}$ conforms to the distribution during the testing phase that corrects a certain amount of error but has a partly over-correction problem.

\subsection{Dataset Construction} \label{data_construction}
\paragraph{Over-corrections obtaining}
The main objective in the data construction phase is generating candidate sentences containing right correction and over-correction for each parallel corpus sample.
Due to data exposure, it is not possible to obtain high-quality over-correction cases by directly inferring on the corpus with a fully trained model.
To address this issue, we propose a data construction method based on k-fold cross inference.
The specific process is shown in Algorithm \ref{algorithm_data}.
Firstly, we randomly divide the training set into K copies.
Subsequently, we use the model obtained by training on k-1 copies to infer partial candidate sentences on the remaining data.
Eventually, with many iterations, we get the candidate sentences of the full training set that correspond to the same distribution as the testing phase.
Specifically, for the FCGEC dataset, we find that setting k to 4 already achieves good results.

\paragraph{Gold Labels Merging}
With k-fold cross inference, we ensure that the model always infers on data not used for training.
This allows us to obtain the same distribution of over-correction as in the test phase, but at the same time doesn't guarantee that the model corrects all errors.
Because the training goal is correct sentences, if there is missing error correction in the candidate sentence it will make the rewriting model still have to learn a part of the error correction task.
In order to be able to completely decouple the two tasks of error correction and rewriting, we add correct corrections to the candidate sentences through MaxMatch \cite{dahlmeier2012better} (M2)  labels, so that the rewriting model only needs to complete the task of filtering over corrections and correct corrections in the candidate sentences.
Specifically, we integrate the M2 labels of the candidate and gold sentences, prioritize the labels of the gold sentence when indexing conflicts, and collaborate the final merged set of M2 labels to the original sentence to obtain the final candidate sentence.

\paragraph{Inference stage}
In the inference phase, we directly use the real output of the error correction system as candidate sentences.
We expect that the LM-Combiner trained on the above data can compare the original and candidate sentences to filter over corrections and retain right corrections.

\begin{algorithm}[t]
	\caption{K-fold Cross Inference}
	\label{algorithm_data}
	\KwIn{$D=\{(x_1, y_1), ... ,(x_n, y_n)\}$, where $x_i$ is the original sentence with the error, $y_i$ is the corrected sentence. The hyper parameter $K$.}
	\KwOut{$D_{candi}=\{(x_1, z_1, y_1), ... ,(x_n, z_n, y_n)\}$, where $z_i$ is the candidate sentence that contains corrective modifications and over-corrections.}
	\BlankLine
	$D_{candi} \leftarrow \{\}$; \\
        
        Randomly divide $D$ into $K$ copies $D_{Split}$;\\
        \ForEach{$D_i$ in $D_{Split}$}{
            $D_{train}$ = $D - D_i$; \\
            Train on $D_{train}$ to get the model $\theta_i$; \\
            Obtain the inference result $Z_i$ of the model $\theta_i$ on $D_i$; \\
            Merge $Z_i$ as candidate sentences with $D_i$ to get $D_{merge}$; \\
	    $D_{candi} = D_{candi} \cup D_{merge}$; \\
        }
        Return $D_{candi}$;
\end{algorithm}

\section{Experiment}

\subsection{Settings}
\paragraph{Dataset}
Restricted by the lack of data, previous CGEC tasks mainly use labeled datasets collected from Chinese as a Foreign Language (CFL) learner sources.
However, \citet{tang2023pre} have discovered by way of human inspection that there is a distributional inconsistency between CFL corpus labeling distributions and native speakers, which may lead to unrealistic metrics.
In recent years, more and more scholars have been working on the construction of CGEC datasets for native speaker corpora, \citetlanguageresource{FCGEC} provide a large-scale multi-reference corpus named FCGEC sourced from native speakers.
Compared to CFL, the CGEC dataset from native speakers is more standardized and has higher annotation quality, but also includes more complex grammatical errors.
We adopt the FCGEC dataset as the main dataset for our experiments, which contains 36,340 sentences of training data, 2,000 sentences of validation set, and 3,000 sentences of test set.

\paragraph{Evaluation metrics}
We follow \citet{zhang2022mucgec}'s setup by using character-level edit metrics to measure the error correction performance of each model.
For the validation set experiments, we use the official evaluation tool ChERRANT
\footnote{\href{https://github.com/HillZhang1999/MuCGEC/tree/main/scorers/ChERRANT}{ChERRANT} is a Chinese GEC evaluation tool that refers to ERRANT, the mainstream GEC evaluation tool in English.}
to evaluate the model based on correction span's P/R/F0.5.
As for the test set, we obtain the same evaluation metrics by submitting the system results in 
CodaLab
\footnote{\href{https://codalab.lisn.upsaclay.fr/competitions/8020}{https://codalab.lisn.upsaclay.fr/competitions/8020}}
online platform.

\paragraph{Model selection}
Reference to mainstream methods of CGEC, our main experiment adopts the model of Bart \cite{lewis2019bart} and GPT2 \cite{radford2019language} architectures as the backbone network.
We use the Chinese Bart model trained by \citet{shao2021cpt} and the series of Chinese GPT2 models trained by \citet{zhao2019uer, zhao2023tencentpretrain} to obtain a good performance on the CGEC task.
Referring to other related work based on the Seq2Seq model \cite{zhang2022mucgec, li2023templategec}, we chose Bart-Large and the equivalent scaled GPT2-medium as the backbone in the main experiment in order to make a fair comparison, and the LM-Combiner also uses the same settings as the GPT2 baseline.


\paragraph{Model hyperparameters}
As a general optimization method, in order to compare the enhancement effect more intuitively, we don't employ some common training techniques in the GEC field (e.g., Src-drop \cite{junczys2018approaching}, label-smoothing \cite{szegedy2016rethinking}, etc.) in the model training phase.
For both models, we use the AdamW \cite{loshchilov2017decoupled} optimizer with 5e-5 learning rate, and 32 batch size for training.
We use the polynomial strategy as a warm-up strategy for learning rate.
Considering the difference in model architectures, the maximum sentence lengths of the Bart and GPT2 models are 256 and 512.
In the testing phase, both generative models inference using beam search with a beam size of 4.

\subsection{Baseline Approaches}
We select several common methods with Seq2Edit and Seq2Seq architectures as baseline models, and pick the one with the largest recall as the system output to validate the effectiveness of the rewriting model.
Our adoption of Chinese GEC model is largely referenced by \citet{zhang2022mucgec, xu2022fcgec}'s related work.
\begin{itemize}
\item \textbf{LaserTagger} \cite{malmi2019encode} is a text generation method based on editing operations that improves the inference speed and reduces the data requirements of the model for the text generation task.
\item \textbf{PIE} \cite{awasthi2019parallel} leverages the power of pre-trained models to efficiently correct grammatical errors through iterative edit tag prediction.
\item \textbf{GECToR} \cite{omelianchuk2020gector} further refines the custom token-level edit tags to map more diverse errors.
\item \textbf{STG} \cite{xu2022fcgec} completes complex grammatical error correction by pipelining three self-encoding models, Switch, Tagger, and Generator, and achieves the SOTA on the FCGEC dataset by jointly training three models.
\item \textbf{Bart} \cite{lewis2019bart,zhang2022mucgec} model has achieved good results on the CGEC task with its denoising pre-training task, and can be used as a representative of the Seq2Seq model.
\item \textbf{GPT2} \cite{radford2019language} model is typically used for generative tasks, and we implemente a GPT model for CGEC as a baseline model following the methodology of Section \ref{baseline}.
\end{itemize}
As a post-processing method, our rewriting model can also be understood as an ensemble of the original sentence and the output of a single system.
Although a single model can't be integrated using traditional voting ensemble methods, the fine-grained PPL-based model ensemble method proposed by \citet{tang2023pre} can still be used as a baseline model for post-processing methods.
Specifically, we replicate three different granularity ensemble approaches based on the same scale of GPT2.
\begin{itemize}
    \item \textbf{Sentence-level} makes a judgment directly from the PPL of the original and output sentences, and only retains sentences with lower perplexity.
    \item \textbf{Edit-level} makes a judgment based on the impact of each editing operation on the PPL of the original sentence, and retains only those operations that reduce the PPL of the original sentence.
    \item \textbf{Edit-combination} permutes all the editing operations and selects the sentence with the lowest PPL among them as the final output as shown in Figure \ref{compare}.
\end{itemize}

\begin{table}[t]
    \centering
    \scalebox{0.90}{
    \begin{tabular}{lccc}
    \toprule
         \multirow{2}{*}{Method} & \multicolumn{3}{c}{FCGEC-test} \\
         \cmidrule(lr){2-4}
        & $P$ & $R$ & $F_{0.5}$\\
        \midrule
        LaserTagger$^*$  & 36.60 & 31.16 & 35.36\\
        PIE$^*$ & 29.15 & 29.77 & 29.27\\
        GECToR (Chinese)$^*$ & 30.68 & 21.65 & 28.32\\
        STG$^*$  & 48.19 & 37.14 & 45.48\\
        \midrule
        Bart-Chinese-large & 37.49 & 38.87 & 37.76 \\
        GPT2-medium & 56.71 & 24.79 & 45.10\\
        \midrule
        Bart-Chinese-large & 37.49 & 38.87 & 37.76 \\
        + Sentence-level & 55.26 & 20.23 & 41.04\\
        + Edit-level & \textbf{58.22} & 24.12 & 45.39\\
        + Edit-combination & 58.16 & 25.63 & 46.38\\
        + LM-Combiner (Ours) & 55.67 & \textbf{39.04} & \textbf{51.30}\\
    \bottomrule
    \end{tabular}
    }
    \caption{
    Experimental results of our method on the FCGEC test set.
    Results with * are reported from the original paper \cite{xu2022fcgec}.
    The first group indicates common Seq2Edit models, the second group indicates Seq2Seq models, and in the last group we choose the highest recall Bart model as a baseline and list some LM-based post-processing methods.    
    }
    \label{main_experiment_table}
\end{table}

\subsection{Main Results}
Table \ref{main_experiment_table} shows the comparison of the performance among different models on the FCGEC test set.
In order to maximize the verification of the performance of the LM-Combiner, we chose the output of the highest recall Bart model as the rewriting input.
As shown in the table, through the rewriting of our LM-Combiner model, we make the output of the original error-correction system substantially improve the precision by 18.2 points while the recall remains basically unchanged, and the $F_{0.5}$ metric improves by 5.8 points compared to the SOTA model.

\begin{figure*}[!ht]
\begin{center}
\includegraphics[scale=0.33]{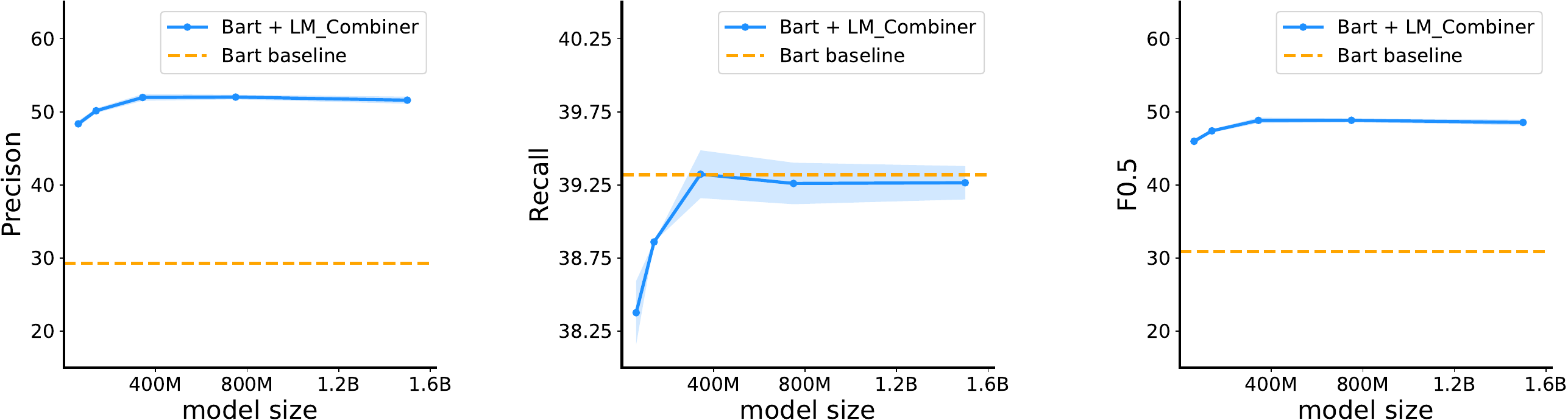} 
\caption{The effect of model size for LM-Combiner on FCGEC valid.
The Bart baseline is the system metric without LM-Combiner rewriting.
For a more accurate evaluation, we average the results of 5 experiments for each size of the model, and the floating part of the figure shows the standard deviation of the metrics.
}
\label{model scale}
\end{center}
\end{figure*}

\begin{figure*}[!ht]
\begin{center}
\includegraphics[scale=0.35]{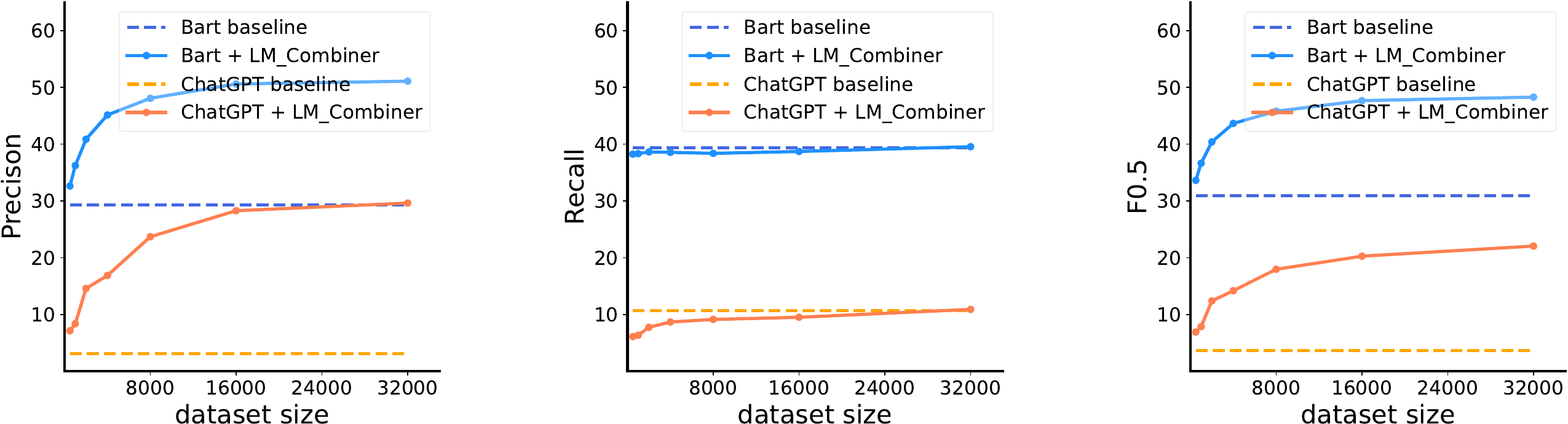} 
\caption{The effect of training dataset size for LM-Combiner on FCGEC valid.
The baseline method represents the metrics for each system without the use of LM-Combiner.
}
\label{data scale}
\end{center}
\end{figure*}

Compared to the PPL-based methods, LM-Combiner does better in recall retention due to the fine over-correction dataset construction.
Although both using GPT2 as the backbone network, the PPL-based approach suffers from the problem of inconsistency between the domains of the pre-trained corpus and the target corpus, which makes the model filter out too much right correction in the re-ranking phase.
Conversely, by constructing a real over-correction dataset under the target domain for training, the LM-Combiner is able to better learn the relationship between over-correction and right correction in the target domain, and thus improves the precision with essentially no decrease in recall.

\section{Analysis}
In this section we will validate and analyze the details of the LM-Combiner through experiments.
\subsection{Effect of Model Scale}
By decoupling the CGEC task, the LM-Combiner model only needs to complete the rewriting task without considering the performance of error correction.
For the simpler rewriting task, we wonder if its rewriting performance is strongly correlated with the scale of the model, thus we use five scales of GPT2, small, base, medium, large, and xlarge, respectively, as the backbone network of the LM-Combiner for the experiments.

As shown in Figure \ref{model scale}, all scales of rewriting model can relatively improve the precision of the GEC system.
The reduction in error recall from rewriting the model becomes smaller and smaller as the model size increases.
In addition to this, we can find that a small-level 62M model can still improve precision by about 18 points compared to the baseline model and essentially preserve the recall of the original system.
For the insignificant change in rewriting performance with model scale, we analyse that this is because the difficulty of the decoupled rewriting task is lower compared to the error correction task, which makes it possible for small models to perform well.

\subsection{Effect of Data Quantity} According to the data construction method in Section \ref{data_construction}, we have obtained the over-correction training set totaling 36,340 sentences of the entire FCGEC training data.
However, in practice it is still a large number in a new domain.
We want to know what amount of parallel corpus will enable us to train a rewriting model works reasonably well through data construction.
Thus, we randomly sample subsets of different sizes from the constructed training set to validate the effect of rewriting the model.

Besides that, \citet{li2023effectiveness,fang2023chatgpt} have evaluated the effectiveness of LLMs (e.g., ChatGPT) on the CGEC task, and the experiments show that there is also a large amount of over-correction in LLMs using the zero-shot and few-shot methods.
Therefore, we follow \citet{fang2023chatgpt}'s approach and also obtain the results of the ChatGPT model on FCFEC for rewriting the model's training as a way to validate the ability of the LM-Combiner for black-box correction systems.
Since there is no data leakage, we directly use the error correction results of ChatGPT as candidate sentences instead of the cross inference method in Section \ref{data_construction}.

The experimental results are shown in Figure \ref{data scale}, LM-Combiner trained at all scales of data amounts is able to alleviate the over-correction problem of the original system to varying degrees.
In particular, thousands of domain training corpora are sufficient to obtain a rewriting model that performs well, both for Bart model and the ChatGPT.
Consistent with \citet{li2023effectiveness}'s evaluation, ChatGPT doesn't perform well on the native speaker CGEC task, with metrics even lower than the Bart baseline model.
Nevertheless, LM-Combiner can still be considered as a low-cost post-processing model, which can effectively relieve over-correction of various GEC systems (including the black-box ChatGPT) on domain-specific datasets.

\subsection{Importance of Gold Labels Merging}
As described in Section \ref{data_construction}, after acquiring the overcorrected data, we merge the gold labels with the overcorrected labels based on the M2 labels as a way to completely decouple the error correction task.
To verify the effect of label merging, we conducted experiments on the original training set and the training set with gold labels merging, respectively.
The experimental results are shown in Table \ref{merge}, where the gold label merging enables LM-Combiner to learn the rewriting task better and retain a higher recall.
It can be said that fully decoupling correction and rewriting tasks by gold labels merging is the key for LM-Combiner to maintain high recall.

\begin{table}[t]
    \centering
    \begin{tabular}{lccc}
    \toprule
             \multirow{2}{*}{Method} & \multicolumn{3}{c}{FCGEC-valid} \\
         \cmidrule(lr){2-4}
            & $P$ & $R$ & $F_{0.5}$\\
    \midrule
        Bart-Chinese-large & 29.31 & 39.32 & 30.88\\
         +LM-C wo merging & \textbf{54.02} & 36.07 & 49.13\\                       
         +LM-C w merging & 53.56 & \textbf{39.25} & \textbf{49.92} \\       
    \bottomrule
    \end{tabular}
    \caption{
    Experimental results on the effectiveness of gold label merging.
    LM-C represents the LM-Combiner model, and merging represents the gold labels merging operation.
    }
    \label{merge}
\end{table}

\section{Related Work}
Compared to the English GEC, the Chinese GEC is just getting started \cite{tang2023pre}.
Early CGEC tasks are mainly researched in the field of non-native language learning, which has a large error rate, and many CFL datasets such as Lang8, CGED \cite{rao2020overview}, and NLPCC18 \citelanguageresource{NLPCC18} are proposed.
On this basis, \citetlanguageresource{MuCGEC} sample and organise the annotation of several CFL datasets, correct the existing annotation problems in them, and propose the MuCGEC dataset with multi-source references.
Recently, more and more scholars \citelanguageresource{FCGEC,NaCGEC} have noticed the problems with CFL datasets and propose a series of datasets based on native speakers' grammatical errors, posing a greater challenge to the CGEC task.

The CGEC task has received increasing attention in recent years.
Responding to the lack of data, \citet{zhao2020maskgec} propose a dynamic mask strategy for data augmentation and improve the robustness of the model.
\citet{yue2022improving} generate high-quality grammatical errors to complete the data augmentation by conditional non-autoregressive error generation model.
In terms of model architecture, \citet{zhang2022syngec} extract the syntactic hidden representation by graph convolutional neural network and incorporate the syntactic information into the GEC system to further improve the error correction performance.
\citet{li2023templategec} fuse the models of the two paradigms in the form of templates and improve the precision of the Seq2Seq model with the help of Seq2Edit model through the detection and correction framework.

Previous researchers have also attempted to explore the potential of causal LMs in GEC tasks.
\citet{yasunaga2021lm} determine the grammatical correctness of a sentence with the help of the PPL of PLMs, and implements a unsupervised GEC framework by assuming that the sentence with the smallest perplexity within a particular set is the correct sentence.
Similarly, \citet{tang2023pre} use the PPL of pre-trained models as a model ensemble method to re-rank the outputs of multiple models.

The large language model represented by ChatGPT \cite{ouyang2022training} is developing rapidly, and there have been some recent related evaluation work \cite{li2023effectiveness, fang2023chatgpt} on LLM on CGEC tasks.
The results indicate that LLM suffers from serious over-correction problems.
Recently \citet{vernikos2023small} use the T5 model for soft aggregation of multiple outputs from LLM, but there are still some common problems of ensemble methods.
In view of this, LM-Combiner is a good solution to alleviate the over-correction problem by directly rewriting individual system outputs without the need for model ensemble.

\section{Conclusion}
In this paper, we propose LM-Combiner, a general rewriting model based on a causal language model, capable of mitigating the problem of over-correction based on the original sentences and single system outputs.
We also propose k-fold cross inference to enable the construction of domain-specific over-correction data for LM-Combiner training.
Experiments show that the proposed method can effectively improve the system precision while ensuring  the recall rate, and it provides a low-cost over-correction solution for existing GEC systems.

\section{Acknowledgement}
We gratefully acknowledge the support of the National Natural Science Foundation of China (NSFC) via grant 62236004 and 62206078.

\nocite{*}
\section{Bibliographical References}\label{sec:reference}

\bibliographystyle{lrec-coling2024-natbib}
\bibliography{lrec-coling2024-example}

\section{Language Resource References}
\label{lr:ref}
\bibliographystylelanguageresource{lrec-coling2024-natbib}
\bibliographylanguageresource{languageresource}

\end{document}